\documentclass[conference]{IEEEtran}
\IEEEoverridecommandlockouts

\usepackage{cite}
\usepackage{booktabs}
\usepackage{multirow}
\usepackage{amsmath,amssymb,amsfonts}
\usepackage{algorithmic}
\usepackage{graphicx}
\usepackage{textcomp}
\usepackage{xcolor}
\usepackage{orcidlink}
\def\BibTeX{{\rm B\kern-.05em{\sc i\kern-.025em b}\kern-.08em
    T\kern-.1667em\lower.7ex\hbox{E}\kern-.125emX}}
    \makeatletter
\newcommand{\linebreakand}{%
  \end{@IEEEauthorhalign}
  \hfill\mbox{}\par
  \mbox{}\hfill\begin{@IEEEauthorhalign}
}
\makeatother
\begin{document}

\title{Taking Flight with Dialogue: Enabling Natural Language Control for PX4-based Drone Agent}

 \author{

 \IEEEauthorblockN{Shoon Kit Lim \orcidlink{0009-0002-8436-7653}}
 \IEEEauthorblockA{
 \textit{University of Southampton Malaysia,}\\
 \textit{Iskandar Puteri, Johor, Malaysia}\\
 skl1g14@soton.ac.uk}
 \and
 \IEEEauthorblockN{Melissa Jia Ying Chong \orcidlink{0009-0002-1225-9303}}
 \IEEEauthorblockA{
 \textit{University of Southampton Malaysia,}\\
 \textit{Iskandar Puteri, Johor, Malaysia}\\
 m.j.y.chong@soton.ac.uk}
 \linebreakand

 \IEEEauthorblockN
 {Jing Huey Khor \orcidlink{10000-0003-2630-0266}}
 \IEEEauthorblockA{\textit{Connected Intelligence Research Group,} \\
 \textit{University of Southampton Malaysia,}\\
 \textit{Iskandar Puteri, Johor, Malaysia}\\
 j.khor@soton.ac.uk}
 \and
 \IEEEauthorblockN{Ting Yang Ling \orcidlink{0000-0003-0914-1072}}
 \IEEEauthorblockA{\textit{Sustainable Electronic Technologies,} \\
 \textit{University of Southampton,} \\
 \textit{Southampton SO17 1BJ U.K.} \\
 ivan.ling@soton.ac.uk}
}

\maketitle

\begin{figure*}
    \centering
    \includegraphics[width=1.0\linewidth]{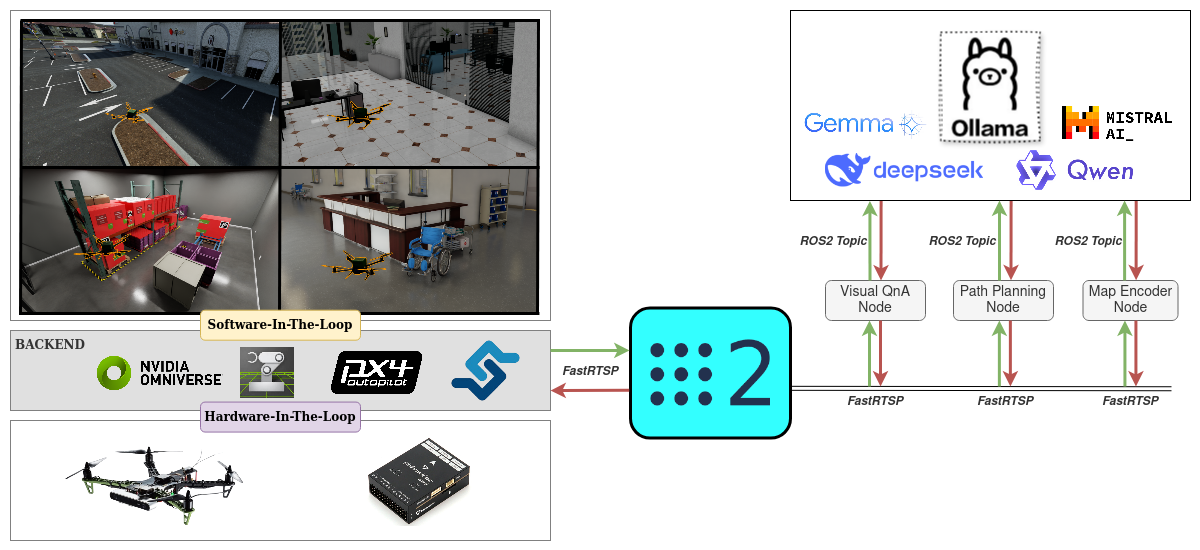}
    \caption{Proposed framework for natural language control of PX4-based drone agents. }
    \label{fig:architecture}
\end{figure*}

\begin{abstract}
Recent advances in agentic and physical Artificial Intelligence (AI) have largely focused on ground-based platforms—such as humanoid and wheeled robots—leaving aerial robots relatively underexplored. At the same time, state-of-the-art UAV multimodal vision-language systems typically depend on closed-source models accessible only to well-resourced organizations. To democratize natural language control of autonomous drones, an open-source agentic framework is presented that integrates PX4-based flight control, Robot Operating System~2 (ROS2) middleware, and locally hosted models using Ollama. Performance is evaluated both in simulation and on a custom quadcopter platform, benchmarking four Large Language Model (LLM) families for command generation and three Vision Language Model (VLM) families for scene understanding. Results indicate that the LLMs, specifically Gemma3, Qwen2.5, and \mbox{Llama-3.2}, consistently produced 100\% valid flight commands, while DeepSeek-LLM demonstrated significantly lower performance at 38\%. Additionally, all VLMs assessed, including Gemma3, Llama3.2-Vision, and Llava1.6, are able to detect the presence of specified objects and give valid binary responses ranging from 97\% to 100\%. The mission success rates varied depending on the model pairings, with the highest success rate of 40\% observed in the Gemma3 LLM and VLM combination. Source code, model configurations, and prompt templates are publicly available at \url{https://github.com/limshoonkit/ros2-agent-ws}. 
\end{abstract}


\section{Introduction}

Industry 5.0 marks a major transformation in manufacturing, moving beyond digital automation to a collaborative approach in which humans and intelligent machines work together\cite{intro_ir5.0}. This paradigm shift has been rapidly gaining traction in various sectors since it was introduced by the European Union in 2021. It pushes the boundaries of automation by emphasizing collaborative environments where humans and robotic agents work side by side to optimize production processes in smart factories.

Recent advances in accelerated computing and machine learning have empowered robots to achieve greater autonomy and more flexible control, thus unlocking more intuitive interaction with humans and their surroundings \cite{intro_llmsforuav}. A key challenge in achieving this lies in equipping robots with the ability to perceive, reason, and act on instructions that mirror human communication. To address that, the Open X-Embodiment project \cite{o2024open} was introduced to enhance robotic learning through diverse, large-scale datasets gathered from various robotic embodiments and environments. This initiative consolidates data from 60 existing robot datasets across 34 research labs, creating a standardized dataset comprising more than 1 million robot trajectories from 22 distinct robot embodiments across 21 institutions to advance robot learning in the cross-embodiment.

At the core of these technological advancements is the human-like language comprehension of Large Language Models (LLMs), particularly Generative Pre-trained Transformers (GPTs) \cite{intro_lvlmforautonomousvehicles}. By leveraging LLMs, users can communicate complex instructions naturally and intuitively, eliminating the need for rigidly coded commands and low-level instruction sets. Building upon this foundation, Vision-Language Models (VLMs) usher in multi-modal intelligence by unifying visual and linguistic understanding. Trained on vast datasets of paired images and text, VLMs can develop a rich semantic understanding of the world. As a result, they excel in vision-language tasks such as image captioning, visual question answering, and complex scene interpretation \cite{intro_uavsmeetllms}.

While LLMs and VLMs have demonstrated generalization and adaptation capabilities across various domains, as evidenced in \cite{doshi24-crossformer} and \cite{huang23vlmaps}, their effectiveness in the context of autonomous navigation remains an open research question. Unmanned Aerial Vehicles (UAVs) face operational trials that differ substantially from those encountered by ground-based robotic systems, such as self-driving cars, robotic manipulators, multi-legged robots, and humanoid robots. These challenges primarily stem from increased motion-induced image degradation, temporal misalignment between image acquisition and processing, and the stringent real-time control requirements of UAVs which further complicates safe human-machine collaboration. 

This paper proposes an embodied agent framework designed for natural language control of PX4-based UAVs, built on top of ros-agents \footnote{\url{https://github.com/automatika-robotics/ros-agents}}. The framework blends NVIDIA Isaac Sim with preset environments, including an outdoor car park, co-working spaces, hospitals, warehouses, and data centers, for Software-In-The-Loop (SITL) simulation. A Robot Operating System~2 (ROS2) wrapper encapsulates Ollama to serve different LLMs and VLMs, from families such as Gemma, DeepSeek, Qwen, and Llama.  Modular tasks are managed by individual nodes, which consist of a visual question-answering node that processes paired images and user queries to generate textual responses, a path planning node that converts a goal point and the current pose into a collision-free trajectory using low-level PX4 flight actions, and a map encoder node that embeds the current pose and semantic information from images into text tokens for map representation. The overall framework is illustrated in Fig. \ref{fig:architecture}.

To summarize, the contributions of this paper are as follows:

\begin{enumerate}
    \item A ROS-based agentic framework that connects the PX4 flight control stack with the Ollama platform.       
    \item A comparative analysis of open-source LLMs and VLMs on Ollama for aerial navigation, evaluated through both simulation (Isaac Sim) and real-world experiments on a custom quadcopter.
\end{enumerate}

The remainder of this paper is structured as follows. Section \ref{sec:RW} provides an overview of related work in this area. In Section \ref{sec:method}, the experimental setup is described, and in Section \ref{sec:results}, the evaluation result and relevant discussion of the embodied agent in guiding a drone to search and approach an object is presented. Finally, Section \ref{sec:conclusion} concludes this work, including a review of the potential future work.

\section{Related Work} 
\label{sec:RW}
Prior robotics research faced limitations in the breadth of tasks addressed and the ability to generalize across diverse embodiments. Although amalgamating language conditioning into deep learning models shows promise, many existing datasets remain specific to particular robot platforms. For example, RT-1 \cite{rt1} employed Transformer architectures and imitation learning, whereas RT-2 \cite{rt2} extended this approach by incorporating web-scale pre-training that represented actions as text tokens, unifying them with natural language tokens in a single Vision–Language–Action (VLA) modeling framework. In parallel, SayCan \cite{saycan} adopted a language model to interpret user instructions and generated a sequence of feasible steps for mobile manipulators, ensuring each step was executable in real-world settings and enabling the robot to complete given high-level human tasks. Meanwhile, SayTap \cite{saytap} put forward a novel interface for language-based quadruped control, building on earlier techniques for language-based modulation.

Several studies have developed novel datasets and benchmarks tailored for LLMs and VLMs in UAV applications, including AerialVLN \cite{aerialvln}, CityNav \cite{citynav}, AeroVerse \cite{aeroverseuavagentbenchmarksuite}, AVDN \cite{avdn}, and UAV-Need-Help \cite{uav-need-help}. These datasets aim to capture realistic aerial dynamics by incorporating continuous action spaces, urban and outdoor environments, and complex language instructions resembling dialogues. However, a recurring criticism in the literature is the lack of validation in sim-to-real deployments. Additionally, challenges remain in addressing the inherent ambiguity and variability of natural language instructions. Some datasets also consider large-scale areas that may not be practical, given UAVs’ limited battery life and operational range. Furthermore, there is insufficient consideration of cluttered environments with noisy sensor data.

The notable success of OpenAI's ChatGPT has catalyzed development into natural language-driven UAV orchestration. Colosseum (formerly Microsoft AirSim) demonstrated this by incorporating ChatGPT with the PX4 autopilot within Unreal Engine 5 \cite{airsim_chatgpt}. This system streamlines photorealistic PX4 SITL simulation by converting a sequence of user prompts into flight commands through MAVLink messages. Similarly, a study utilizing the Gazebo simulator showcased ChatGPT generating PX4 commander commands for rudimentary functions, for instance arming, disarming, takeoff, landing, and flight mode transitions \cite{fromwordstoflight}. Expanding to the broader ROS ecosystem, ROSA \cite{rosa} exploited LLMs for general robotic control through a reasoning-action-observation loop, granting it capabilities to interpret natural language, plan and execute actions via dynamically callable tools, and iteratively refine responses until mission resolution. 
This work diverges from existing approaches by avoiding sole reliance on the closed-source OpenAI ChatGPT as the LLM backbone. Instead, it utilizes Ollama as the LLM and VLM serving platform, providing flexibility to integrate a broader range of models.
\section{Methodology}
\label{sec:method}

\begin{figure}
    \centering
    \includegraphics[width=1.0\linewidth]{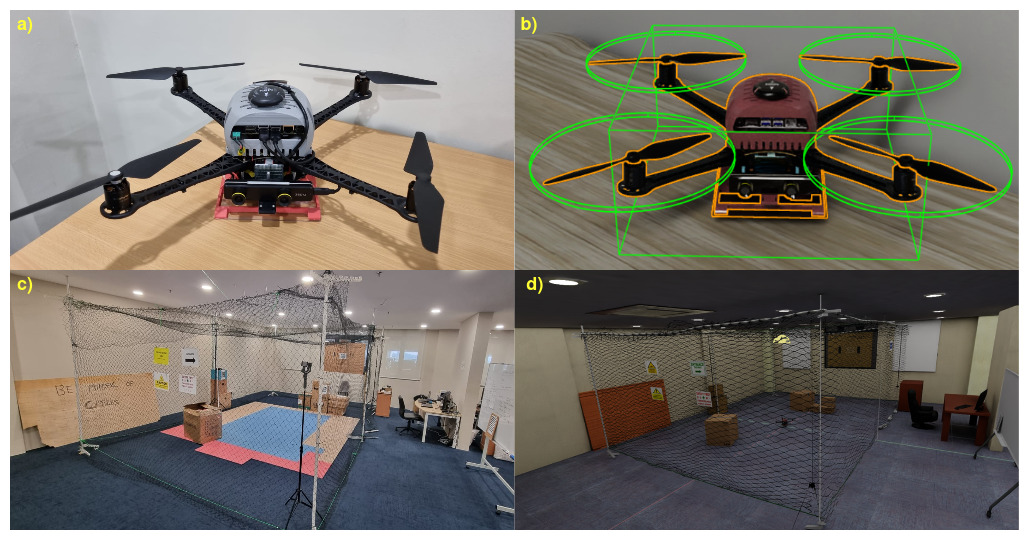}
    \caption{Experimental setup showing (a) a customized quadcopter, (b) the digital twin of the quadcopter, (c) an indoor flight environment, and (d) the digital twin of the environment.}
    \label{fig:setup}
\end{figure}

The proposed interface was evaluated and validated in simulated and real-world environments. The experimental setup is depicted in Fig. \ref{fig:setup}, which includes the construction of digital twins for a custom-built quadcopter and the indoor testing environment. The quadcopter is fitted with an NVIDIA Jetson Orin Nano Dev Kit for onboard computation, a ZED Mini camera for visual-inertial odometry (VIO), and a Pixhawk 6c Mini flight controller for motor control. The physical testing area is enclosed with nylon netting measuring 7m $\times$ 4.5m $\times$ 2.2m, with cardboard boxes placed throughout the space to serve as static obstacles.

The SITL simulations were conducted in a virtual environment using Isaac Sim as the rendering and physics engine. These simulations were run on an Ubuntu 22.04 workstation containing an NVIDIA RTX 3080Ti (16 GB) GPU. A separate Ubuntu 22.04 workstation, also featuring an RTX 3080Ti (16 GB) GPU, was used to host the Ollama models whereby it acts as a remote server that provides linguistic reasoning capabilities to the physical and simulated systems through a standard local network connection.

The embodied agent is responsible for guiding a quadcopter from a known initial configuration to a predefined goal location $G$, while adhering to a set of operational constraints. Formally, the agent must generate a discrete-time trajectory:

\begin{figure}
    \centering
    \includegraphics[width=1.0\linewidth]{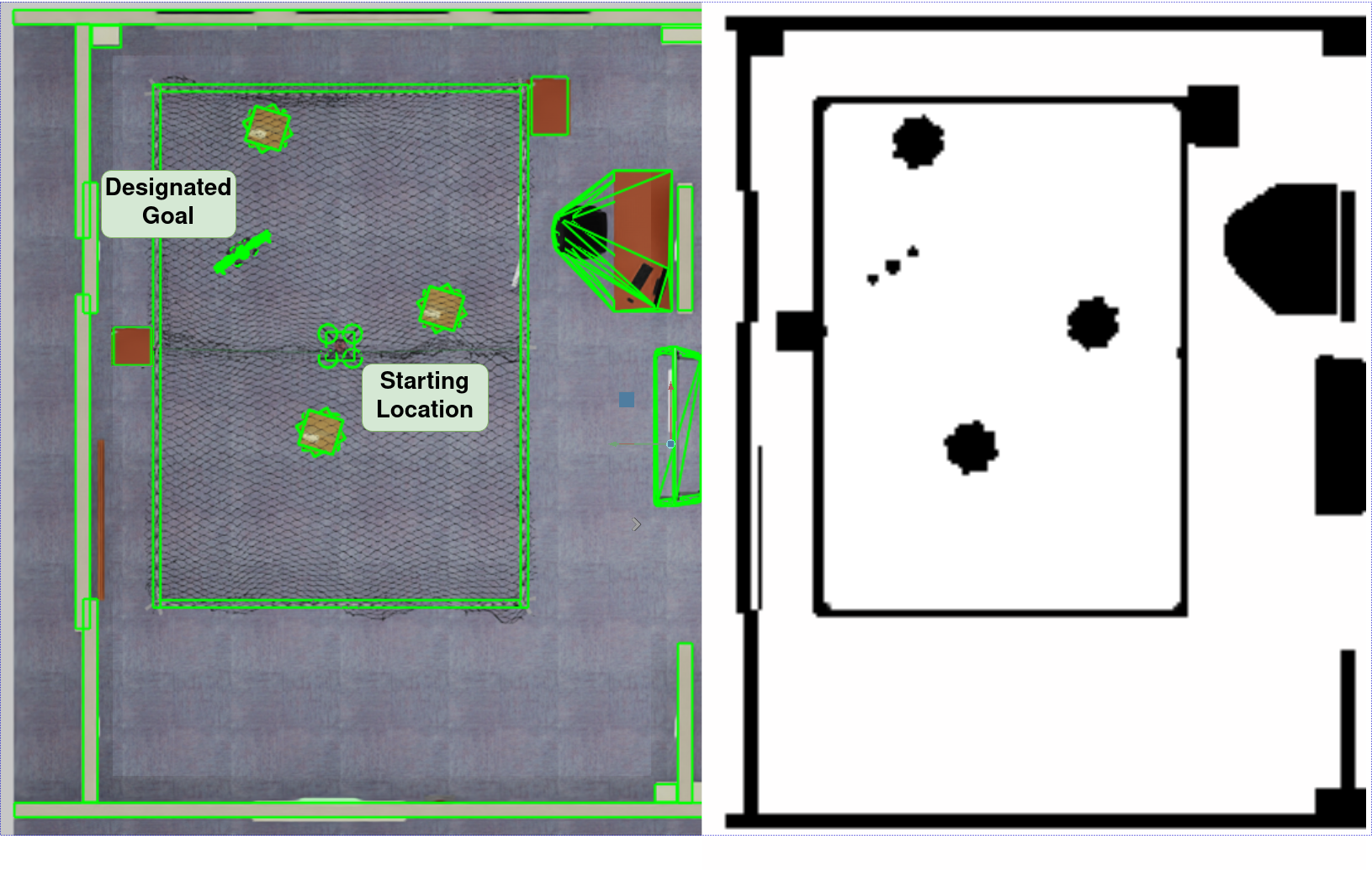}
    \caption{Overview of the quadcopter operational boundary from a top-down perspective with its equivalent 2D occupancy grid map.}
    \label{fig:task}
\end{figure}

\begin{figure*}
    \centering
    \includegraphics[width=1.0\linewidth]{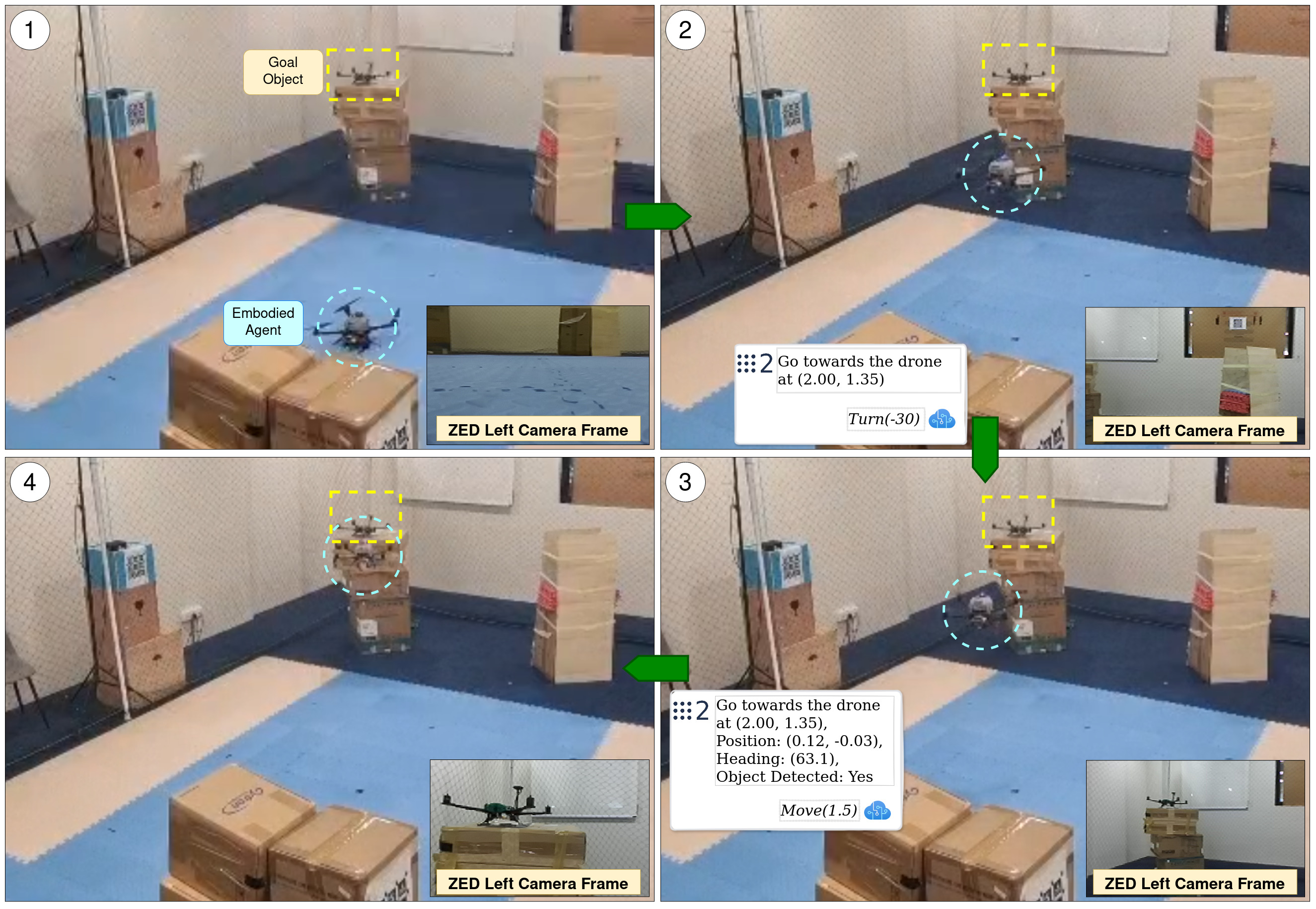}
    \caption{Embodied agent demonstration in real flight.}
    \label{fig:real}
\end{figure*}

\begin{equation}
    \text{Traj}_{0:K} = (S_0, S_1, \dots, S_K)
\label{eq:traj}
\end{equation}

where $S_0$ denotes the known initial state and $S_k$ the desired terminal state. The state transition is determined by the dynamics of the PX4 autopilot:

\begin{equation}
    S_{k+1} = \text{PX4}(S_k, A_k)
\label{eq:state}
\end{equation}

in which $A_k$ is the control input issued by the agent at discrete step $k$.

The action space of the agent is defined by two deterministic motion primitives with restricted parameter ranges:
        \begin{itemize}
            \item If $A_k = \texttt{Turn}(\theta);$, then $\theta \in [\pm90^\circ]$.  Negative values of $\theta$ correspond to the left yaw (counter-clockwise), and positive values correspond to the right yaw (clockwise).
            \item If $A_k = \texttt{Move}(d);$, then $d \in [\pm3.0]$ meters. This represents the forward and backward translation of the quadcopter by a distance $d$ along its body frame's forward axis.
        \end{itemize}

At each time index $k$, the agent computes
\begin{equation}
    A_k = \mathcal{LLM}(C_{k}, \mathcal{H}_k^{(N)}, VLM(\mathcal{V}_k))
\label{eq:action}
\end{equation}
where $C_k$ represents task-specific contextual queries (e.g., state metadata, mission instructions, environmental cues), $\mathcal{H}_k^{(N)}$ denotes the history of the $N$ most recent action commands executed by the agent, $\mathcal{V}_k$ is the current set of visual observations acquired from the quadcopter's onboard cameras. An LLM generates action commands given contextual inputs and recent action history, while a VLM assesses the visual input to determine the presence of mission-relevant objects within the current scene. The LLM and VLM operate collaboratively to achieve the mission objective.

In addition to the action space constraints, the quadcopter's operation is further curbed by its physical boundaries and the mission length. The positional state $S$ of the quadcopter must remain within a predefined bounded three-dimensional spatial region $\mathcal{B} \in \mathbb{R}^3$ throughout the entire mission. Moreover, the agent must not exceed the maximum allowed time step $K_{\max} \in \mathbb{R}^+$:

\begin{equation}
\mathcal{S}[x, y, z] \in \mathcal{B}, \quad t \in [0, K_{\max}]
\label{eq:constraint}
\end{equation}

\begin{table*}[t]
\centering
\caption{Evaluation of Embodied Navigation Agent with different LLMs and VLMs}
\label{tab:evaluate}
\begin{tabular*}{\textwidth}{@{\extracolsep{\fill}}ccccccc}
    \toprule
    LLM Model & Parameter Size & \multicolumn{1}{c}{\begin{tabular}[c]{@{}c@{}}LLM Valid\\Commands (\%)\end{tabular}} & 
    VLM Model & Parameter Size & \multicolumn{1}{c}{\begin{tabular}[c]{@{}c@{}}VLM Valid\\Detections (\%)\end{tabular}} & 
    \multicolumn{1}{c}{\begin{tabular}[c]{@{}c@{}}Mission\\Success Rate (\%)\end{tabular}} \\
    \midrule
    \multirow{3}{*}{Gemma3} & \multirow{3}{*}{4B} & \multirow{3}{*}{100} & Gemma3 & 12B & 100 & \textbf{40} \\
                            &                     &                      & Llama3.2-Vision & 11B & 100 & 30 \\
                            &                     &                      & Llava1.6 & 7B & 98 & 30 \\
    \midrule
    \multirow{3}{*}{Qwen2.5} & \multirow{3}{*}{3B} & \multirow{3}{*}{100} & Gemma3 & 12B & 100 & 30 \\
                             &                     &                      & Llama3.2-Vision & 11B & 100 & 35 \\
                             &                     &                      & Llava1.6 & 7B & 97 & 30 \\
    \midrule
    \multirow{3}{*}{Llama-3.2} & \multirow{3}{*}{3B} & \multirow{3}{*}{100} & Gemma3 & 12B & 100 & 30 \\
                               &                     &                      & Llama3.2-Vision & 11B & 100 & 30 \\
                               &                     &                      & Llava1.6 & 7B & 98 & 35 \\
    \midrule
    \multirow{3}{*}{DeepSeek-LLM} & \multirow{3}{*}{7B} & \multirow{3}{*}{38} & Gemma3 & 12B & 100 & 0 \\
                                  &                     &                      & Llama3.2-Vision & 11B & 100 & 5 \\
                                  &                     &                      & Llava1.6 & 7B & 98 & 0 \\
    \bottomrule
\end{tabular*}
\end{table*}

During each simulation episode, a target object and its location are randomly generated within the quadcopter’s operational boundary, along with the placement of random obstacles, as shown in Fig.~\ref{fig:task}. The quadcopter always start at the same initial location. Obstacles are uniformly set to a height of 1.5 m, with their width and length scaled to match the quadcopter's wingspan of 0.56 m. A minimum clearance of 1 m is maintained between all obstacles and the target.

At the beginning of each experimental episode, the quadcopter is commanded to transition to OFFBOARD mode, take off,  and ascend to a nominal altitude of approximately 1 m. This altitude is preserved during the navigation trial, constraining the vehicle to a fixed vertical plane. Upon reaching the desired height, the embodied agent assumes autonomous control of the aerial platform.

Subsequently, a human operator initiates the mission by specifying a target object to search and a textual description of the coordinate of the goal by publishing a ROS2 string message.  The target specified in the experiment is a visually discernible object, such as another robot or drone. The mission is considered successful if the quadcopter locates and approaches the target within a 0.5 m radius.

Additional conditions that constitute mission failure include collisions with obstacles or violations of operational boundaries, as well as exceeding the maximum permissible execution time. During real-world experiments, a human operator is present as a safety fallback mechanism. If a potential crash is imminent, the operator will switch the flight mode to POSITION mode, causing the quadcopter to hover in place, thus allowing the operator to instantly regain manual control.

\section{Results and Discussion}
\label{sec:results}

This study evaluated the most recent open-source LLMs available at the time of writing, including Google Gemma3, Alibaba Qwen2.5, Meta Llama 3.2, and Deepseek-LLM. Proprietary models requiring paid Application Programming Interface (API) access, such as OpenAI’s ChatGPT-4 and Anthropic’s Claude, were omitted from the analysis. Due to hardware memory limitations, the study could not assess the full spectrum of model configurations, which include varying quantization levels, instruction embeddings, and extended parameter sizes.

To enhance deterministic model outputs and minimize stochastic variability, each model’s internal chain-of-thought reasoning was suppressed via system-level prompting. The sampling temperature for all LLMs and VLMs was also held at a low value of 0.2. The models’ built-in conversational history was disabled as well. Instead, each evaluation trial received a fixed history comprising the five most recent valid commands, which were explicitly supplied as user prompts. Each trial was allotted a maximum of 20 inference steps to complete its assignment. Each combination ran for a total of 20 episodes.

The evaluation process for LLMs primarily focused on their ability to adhere to system-level instructions by generating syntactically correct navigation commands, specifically "Turn" or "Move" with magnitudes that remained within predefined ranges. Any output containing extraneous textual elements, symbols, markdown formatting, or standard AI disclaimers was classified as invalid. Importantly, the LLMs were not evaluated based on the arithmetic precision of angle calculations or movement distance estimations given the quadcopter's current position and heading relative to the goal coordinates.

The performance of the vision-language models was measured by their binary “Yes” or “No” response to whether a given object class appeared in the camera’s field of view. Three object classes were used for validation: humanoid robots, drones, and quadcopters. Object detection correctness was not considered.

A summary of the evaluation results is provided in Table \ref{tab:evaluate}, with the highest mission success rate of 40\% observed for the Gemma LLM-VLM pairing. The results also indicate a strong association between the proportion of valid navigation commands generated by the language model and the corresponding mission completion rate. This relationship is particularly evident in the case of the DeepSeek model, which exhibited a low command validity rate of 38\% and consequently failed to achieve meaningful mission success. Notably, even in scenarios where valid commands were generated and object presence was accurately detected at 100\%, some mission failures occurred due to suboptimal command values, resulting in the quadcopter either undershooting or overshooting the 0.5 m target radius. Furthermore, false positive and false negative object detection are also attributed to the low mission success rate.  Llava1.6 is the only model among the three sampled VLMs that will produce a response outside of "Yes" and "No" across the experiment. Nevertheless, because each model pairing was evaluated on a relatively small number of episodes, the modest differences in success rates observed between the Gemma, Qwen, and Llama LLMs may reflect random variation and therefore may not reach statistical significance.

Qualitative observations during the experiments offer additional insights into the behavior of the selected models. First of all, the LLM component occasionally repeats identical commands with similar values in succession. This behavior is likely due to a low-temperature setting, which discourages exploration. Additionally, the DeepSeek model keeps producing erratic or reasoning-based responses, hence producing invalid commands even though system-level prompting required it to only generate responses in two formats "Move" and "Turn". As for the VLM component, there appears to be a strong correlation between mission success rates and a clear, unobstructed line of sight between the camera viewport and the target object. When the target is occluded by simulated obstacles upon simulation resets, success rates are noticeably reduced. For real-world deployment, only Gemma was chosen for the LLM and VLM components as this combination exhibited the highest mission success rate throughout the simulations. In addition, Gemma has an inherent tendency to output more conservative motion values which reduce the risk of collision. Figure \ref{fig:real} illustrates one of the successful sequences showcasing the interaction of query, response, and actions undertaken by the quadcopter through generated commands.

Several limitations of the current study deserve thorough attention. Firstly, the PX4 flight control system operates within a North-East-Down (NED) coordinate frame. Consequently, explicit transformations to the East-North-Up (ENU) convention, which is commonly employed in robotic systems, are necessary to ensure proper vector alignment of the agent's command and the robotic platform. Secondly, the agent is limited to issuing only forward and yaw rate commands, thereby not fully utilizing the omnidirectional maneuverability typically available in aerial robotic platforms. Additionally, the image data transmitted through ROS is encoded in the RGB8 format, which could introduce potential discrepancies, as some models may have been trained using other color channel formats, such as BGR8, prevalent in the OpenCV library. The current evaluation also lacks a quantitative analysis of several vital system metrics, including end-to-end latency, token usage of the language model, and path optimality, which could be assessed using shortest path algorithms such as A* on a fully observable map. Moreover, there is an absence of ablation studies examining the impact of conversational history or few-shot examples on mission success rates. While domain randomization has been applied to account for sensor and control noise within the simulation environment, there are no synthetic imaging artifacts, such as motion blur or exposure irregularities caused by electronic shutters that may further address the sim-to-real gap.

\section{Conclusion}
\label{sec:conclusion}

To advance realistic UAV control through natural language and multimodal vision understanding, this work presented an agentic framework built on the Ollama serving platform, enabling flexible integration of open-source LLMs and VLMs tailored to specific resource and mission requirements. While challenges remain for real-world deployment, leading open-source models have shown strong potential in scene interpretation, and accurate natural language-based command execution. Incorporation of a dedicated object detection module such as YOLO or DINO may further enhance the VLM detection capabilities. These capabilities are particularly promising for industrial applications such as routine inventory tracking and equipment inspection. Future research may focus on curating domain-specific datasets and applying techniques such as Reinforcement Learning from Human Feedback (RLHF) and knowledge distillation to improve task performance. Additionally, the development of a single end-to-end VLA model with fewer than one billion parameters offers a promising path toward efficient, fully edge-deployable systems with reduced latency and minimal dependence on cloud infrastructure.

\bibliographystyle{IEEEtran}
\bibliography{IEEEabrv,References}


\end{document}